\documentclass[11pt,a4paper]{article}
\usepackage[hyperref]{acl2021}
\usepackage{times}
\usepackage{latexsym}

\usepackage{makecell}
\usepackage{microtype}

\aclfinalcopy 

\usepackage{multirow}
\usepackage{amsmath}
\usepackage{capt-of}
\usepackage{tabularx}
\usepackage[caption=false]{subfig}
\usepackage{epsfig}
\usepackage{amssymb}
\usepackage{amsfonts}
\usepackage{booktabs}
\usepackage{scalerel}
\usepackage[inline]{enumitem}
\usepackage{listings}
\usepackage{varwidth}
\usepackage[export]{adjustbox}
\usepackage{cleveref}

\usepackage{stmaryrd}
\usepackage{bbm}

\newcommand{\sptk}[1]{\texttt{[#1]}}
\newcommand{\eqform}[1]{Equation~(\ref{#1})}

\usepackage{algorithm}
\usepackage[noend]{algpseudocode}

\definecolor{deepblue}{rgb}{0,0,0.5}
\definecolor{officeblue}{RGB}{0,102,204}
\definecolor{deepred}{rgb}{0.6,0,0}
\definecolor{deepgreen}{rgb}{0,0.5,0}
\definecolor{mybrickred}{RGB}{182,50,28}

\definecolor{fillcolor}{RGB}{216,217,252}

\algnewcommand\algorithmicrequireb{{\hspace{0.85cm}}}
\algnewcommand\INPTDESCB{\item[\algorithmicrequireb]}

\algnewcommand\algorithmicfuncdesc{\textbf{Function:}}
\algnewcommand\FUNCDESC{\item[\algorithmicfuncdesc]}
\algnewcommand\algorithmicfuncdescb{{\hspace{1.48cm}}}
\algnewcommand\FUNCDESCB{\item[\algorithmicfuncdescb]}
\algnewcommand{\algorithmicgoto}{\textbf{goto}}
\algnewcommand{\Goto}[1]{\algorithmicgoto~\ref{#1}}

\usepackage{amsmath,amsfonts,bm}

\def\eqref#1{equation~\ref{#1}}

\def\1{\bm{1}}

\DeclareMathAlphabet{\mathsfit}{\encodingdefault}{\sfdefault}{m}{sl}
\SetMathAlphabet{\mathsfit}{bold}{\encodingdefault}{\sfdefault}{bx}{n}

\DeclareMathOperator*{\argmax}{arg\,max}

\newcommand\cnndm{CNN / DailyMail}

\newcommand\stosft{\textit{s2s-ft}}

\title{\stosft{}: Fine-Tuning Pretrained Transformer Encoders \\ for Sequence-to-Sequence Learning}

\author{Hangbo Bao, Li Dong, Wenhui Wang, {Nan Yang}, {Furu Wei}
\\
Microsoft Research \\
\url{https://github.com/microsoft/unilm/tree/master/s2s-ft}
}

\date{}

\begin{document}
\maketitle
\begin{abstract}
Pretrained bidirectional Transformers, such as BERT~\cite{bert}, have achieved significant improvements in a wide variety of language understanding tasks, while it is not straightforward to directly apply them for natural language generation. In this paper, we present a sequence-to-sequence fine-tuning toolkit \stosft{}, which adopts pretrained Transformers for conditional generation tasks. Inspired by UniLM~\cite{unilm,unilmv2}, we implement three sequence-to-sequence fine-tuning algorithms, namely, causal fine-tuning, masked fine-tuning, and pseudo-masked fine-tuning. By leveraging the existing pretrained bidirectional Transformers, experimental results show that \stosft{} achieves strong performance on several benchmarks of abstractive summarization, and question generation. Moreover, we demonstrate that the package \stosft{} supports both monolingual and multilingual NLG tasks.
The \stosft{} toolkit is available at \url{https://github.com/microsoft/unilm/tree/master/s2s-ft}.
\end{abstract}

\begin{figure*}
\begin{center}
\includegraphics[width=\textwidth]{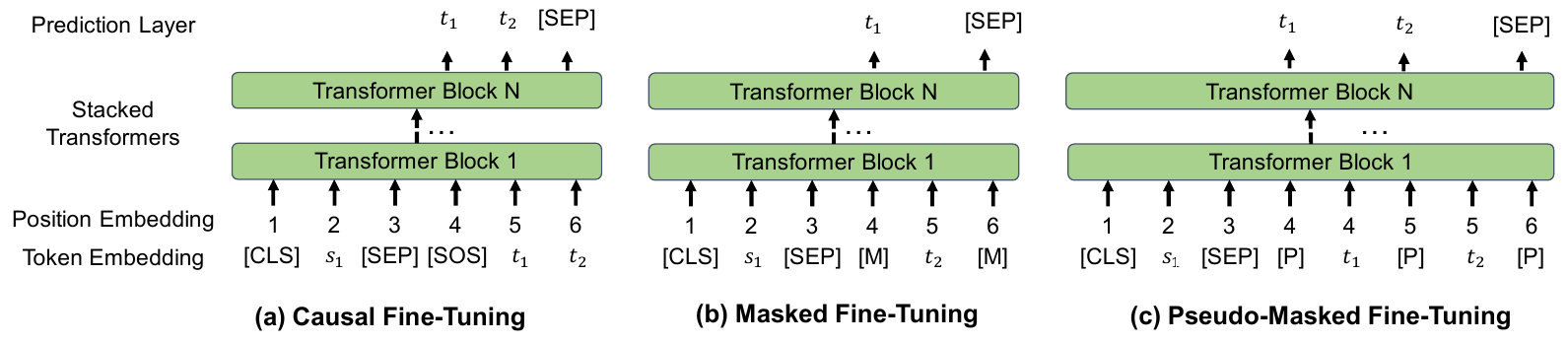}
\caption{Overview of different fine-tuning methods. 
We pack the source and target sequence together to form the input and use specific attention masks shown in Figure~\ref{fig:att:mask} to perform sequence-to-sequence fine-tuning.
\sptk{M} and \sptk{P} denote the masked token, \sptk{CLS} and \sptk{SOS} the start-of-sequence tokens, and \sptk{SEP} the end-of-sequence token.
For causal fine-tuning, each target token is fed into the model in order to predict the next token.
For masked fine-tuning, we randomly mask some tokens in target sequence and train the model as masked language modeling.
For pseudo-masked fine-tuning, we insert a pseudo mask for each target token, and assign them with the same position embeddings.
}
\label{fig:overview}
\end{center}
\end{figure*}

\section{Introduction}
\label{sec:intro}

Pretrained bidirectional Transformers~\citep{bert,xlnet,unilm,roberta,xlmr,electra,unilmv2} have achieved remarkable success on various NLP tasks, such as text classification, and question answering.
The BERT-like models are usually pretrained by the masked language modeling task~\cite{taylor1953cloze,bert}, which learns to predict masked tokens based on given context.
However, due to the bidirectionality nature, it is not straightforward to directly apply the pretrained bidirectional Transformers to language generation tasks~\citep{bertmarkovlm}.

There have been several attempts to achieve the above goal.
\citet{liu-text} use pretrained BERT~\cite{bert} as an encoder, and randomly initialize a Transformer-based decoder with larger learning rate.
\citet{bertshared} initialize the encoder and decoder with different combinations of BERT, GPT~\cite{gpt}, and RoBERTa~\cite{roberta} models.
Despite achieving promising results, the performance is still far behind the jointly pretrained encoder-decoder models, such as BART~\cite{bart} and T5~\cite{t5} on generation tasks.
We argue that the capability of pretrained bidirectional Transformers has not been fully unleashed on sequence-to-sequence tasks.

In this paper, we present a toolkit (named as \stosft{}) used to fine-tune pretrained bidirectional Transformers on conditional language generation tasks, such as abstractive summarizaiton, and question generation.
We follow unified modeling as in~\cite{unilm}, which shares the same Transformer parameters for both encoding and decoding.
Sequence-to-sequence modeling is achieved by employing well-designed self-attention masks in bidirectional Transformers. In other words, the source tokens can attend to each other, while the target tokens can only attend to the left-side context.

We implement three fine-tuning algorithms in \stosft{}.
Firstly, causal fine-tuning introduces a position shift for decoding target sequences as in causal language modeling, so that all the decoding tokens can be trained with one forward pass.
Secondly, masked fine-tuning randomly masks some target tokens and learns to recover them. The method minimizes the mismatch between pre-training and fine-tuning.
Thirdly, pseudo-masked fine-tuning appends pseudo masks into the original target sequence, which combines the benefits of the above two methods.

We build the \stosft{} toolkit upon HuggingFace's Transformers library~\cite{HuggingFacesTS}.
We conduct extensive experiments on several language generation benchmarks, such as XSum and \cnndm{} for abstractive summarization, and SQuAD question generation.
We also compare off-the-shelf pretrained bidirectional Transformers (i.e., BERT~\cite{bert}, RoBERTa~\cite{roberta}, ELECTRA~\cite{electra}, and UniLM~\cite{unilm,unilmv2}) for sequence-to-sequence learning.
In addition, we show that \stosft{} can be easily applied to multilingual language generation tasks by using XLM-RoBERTa~\cite{xlmr} as the multilingual pretrained model.
Experimental results demonstrate that \stosft{} achieves strong performance across different tasks, and languages.

\begin{figure*}
\begin{center}
\includegraphics[width=\textwidth]{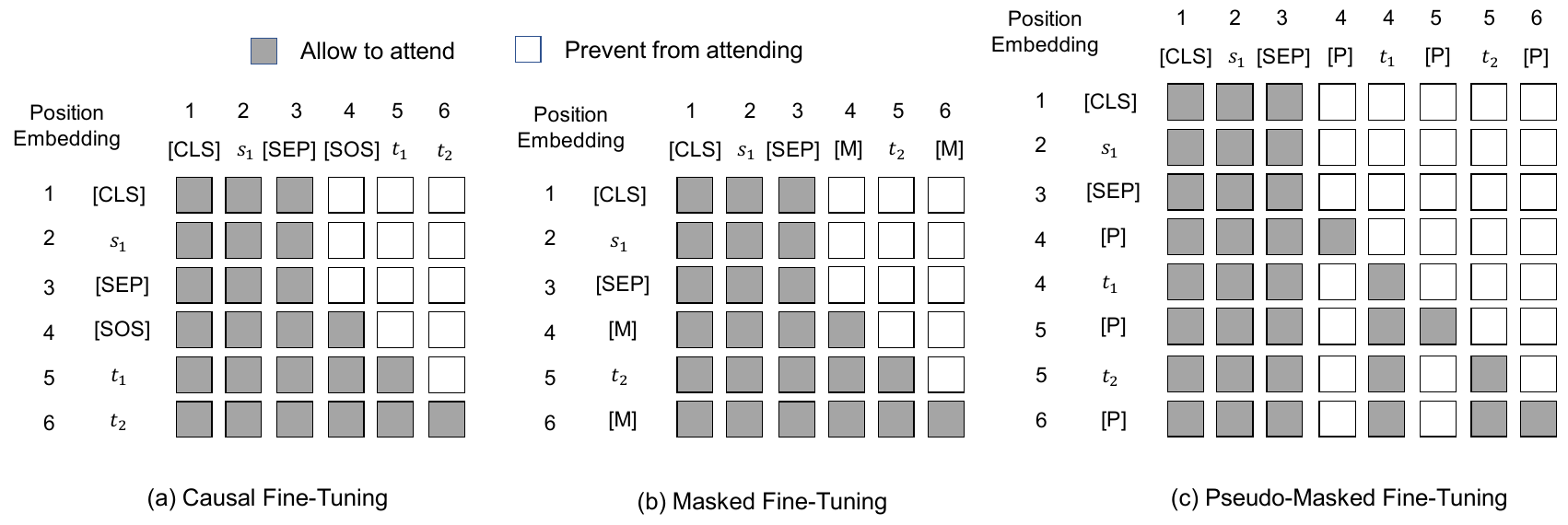}
\caption{Self-attention masks for different fine-tuning methods. 
Tokens in the target sequence can attend to source tokens, left context in the target sequence and itself.
For pseudo-masked fine-tuning, the mask token \sptk{P} can only be attended by itself. 
}
\label{fig:att:mask}
\end{center}
\end{figure*}

\section{Sequence-to-Sequence Fine-Tuning}
\label{sec:method}

Sequence-to-sequence learning aims at generate a target sequence $t = t_1, \cdots , t_{|t|}$ by conditioning on the given source sequence $s = s_1 , \cdots , s_{|s|}$.
In \stosft{}, all tokens are encoded into hidden vectors by Transformer~\cite{transformer}.
The target tokens are autoregressively generated via:
\begin{align}
p(t | s) = \prod_{i=1}^{|t|}{ p(t_i | t_{<i}, s) } \label{eq:seq2seq}
\end{align}
where $t_{<i} = t_1,\cdots,t_{i-1}$.

First, the model is initialized by a pretrained Transformer. Then we use a sequence-to-sequence learning objective to fine-tune the network.
Inspired by UniLM~\cite{unilm}, we share the same model architecture and parameters for both encoding and decoding, which reduces the modeling discrepancies between pre-training and fine-tuning.

Figure~\ref{fig:overview} shows an overview of three sequence-to-sequence fine-tuning algorithms implemented in the \stosft{} toolkit.
We employ special tokens to indicate the boundary of sequences.
For example, \sptk{CLS} is the first source token, and \sptk{SEP} indicates the end of sequences.
Sequence-to-sequence learning is achieved by using well-designed self-attention masks~\cite{unilm}.
As shown in Figure~\ref{fig:att:mask}, all the source tokens can attend to each other, while a target token can only attend to the previously generated tokens and source sequence.
We encode the source sequences as conventional bidirectional Transformers.
The main difference between three fine-tuning methods lies in how to decode target sequences.

\subsection{Causal Fine-Tuning}
\label{sec:causal:ft}

The first method learns to decode target in a similar way as causal language models, such as GPT~\cite{gpt}.
In the decoding part, the model generates the current token by feeding the previous prediction at each time step.
As shown in Figure~\ref{fig:overview}(a), we feed the start-of-sequence token \sptk{SOS} into the model and predict $t_1$ by conditioning on the hidden state.
Similarly, $t_1$ is the input at the next time step, which is used to produce $t_2$.
The target tokens are completely generated until the end-of-sequence token \sptk{SEP} is emitted.

The fine-tuning objective is to maximize the likelihood of generating target tokens conditioning on the source sequence.
Unlike masked language model pre-training, where only a portion of tokens are masked and predicted, causal fine-tuning can gather supervision signals from every target predictions within one forward pass.
However, the method involves a \textit{position shift} between model input and prediction in the decoding part, which results in a discrepancy compared with bidirectional Transformer pre-training.

\subsection{Masked Fine-Tuning}
\label{sec:masked:ft}

Following~\citet{unilm}, we randomly mask a certain percentage of target tokens, and learn to recover them.
The masked fine-tuning algorithm is identical to masked language model pre-training, despite we use a sequence-to-sequence self-attention mask as shown in Figure~\ref{fig:att:mask}(b).
The masked position is supposed to predict the current target token, while other tokens are given as context.
Notice that the end-of-sequence token \sptk{SEP} can also be masked during fine-tuning in order to learn when to terminate the decoding process.

The fine-tuning objective is to maximize the likelihood of masked tokens given source and uncorrupt target tokens.
The method overcomes the position-shift discrepancy between pre-training and fine-tuning described in Section~\ref{sec:causal:ft}.

\subsection{Pseudo-Masked Fine-Tuning}
\label{sec:pseudo:ft}

Following~\citet{unilmv2}, we append pseudo-masked tokens \sptk{P} for all the target tokens.
The pseudo mask is assigned with the same position embedding as the corresponding original token.
Compared with masked fine-tuning, the original tokens are still kept in the input rather than being masked.

The self-attention mask used for pseudo-masked fine-tuning is illustrated in Figure~\ref{fig:att:mask}(c).
All the source tokens can be accessed by others.
The pseudo masks and target tokens can only attend to the previous given tokens and themselves.
Moreover, the original target token instead of its corresponding pseudo mask is attended by the future time steps.

As shown in Figure~\ref{fig:overview}(c), the target tokens are predicted at the positions of pseudo masks.
The fine-tuning objective is to maximize the likelihood of target tokens given source sequence.
Pseudo-masked fine-tuning gets the best of the above two methods.
The algorithm avoids the position-shift discrepancy compared with causal fine-tuning (Section~\ref{sec:causal:ft}).
Moreover, all the target tokens can back-propagate error signals, rather than only a portion of target sequence are masked and predicted as in masked fine-tuning (Section~\ref{sec:masked:ft}).

\subsection{Decoding}
\label{sec:decoding}

Given input source $s$, the target sequence is autoregressively generated via $\hat { t } = \argmax_{t'}{ p( t' | s ) }$, where $p( t' | s )$ is factorized as in~\eqform{eq:seq2seq}.
We approximately find the best decoding results by greedy search or beam search, similar in conventional encoder-decoder methods.

It is worth noting that the hidden states of previous time steps can be cached without re-computing them during the decoding process.
So the decoding process has the same computation complexity compared with conventional Transformer sequence-to-sequence models.
Moreover, the implementation becomes more unified because \stosft{} uses the same architecture for both encoding and decoding.
In contrast, conventional Transformers need to distinguish encoder and decoder~\cite{transformer}, where different architecture are implemented.

For causal fine-tuning, a start-of-sequence token is fed into the model in order to predict the first target token. Then we in turn append the prediction to input and generate the next token. We repeat the process until the end-of-sequence token is emitted.
In contrast, the other two methods use a mask \sptk{M}/\sptk{P} as input to predict the current target token. The mask will be substituted with its prediction in the next time step.

\section{Experiments}
\label{sec:exp}

\stosft{} is built upon HuggingFace's Transformers library~\cite{HuggingFacesTS}, so that we can load various off-the-shelf pretrained models.
We implement the sequence-to-sequence fine-tuning algorithms described in Section~\ref{sec:method}.
We conduct experiments on a set of language generation benchmarks, including abstractive summarizaiton, and question generation.
The hyperparameters are chosen on the development set of each dataset.

\subsection{Benchmarks}

\begin{table}[t]
\centering
\small
\begin{tabular}{lll}
\toprule 
\textbf{Dataset} & \textbf{\#Train/\#Dev/\#Test} & \textbf{Language}  \\ \midrule
\multicolumn{3}{l}{~~\emph{Abstractive Summarization}} \\
CNN / DailyMail & 287k/13k/11k & English \\
XSum & 204k/11k/11k & English \\
Gigaword$_{\text{fr}}$ & 500k/5k/5k & French \\ 
Gigaword$_{\text{zh}}$ & 500k/5k/5k & Chinese \\ \midrule
\multicolumn{2}{l}{~~\emph{Question Generation}} \\
SQuAD & 76k/11k/12k & English \\
WebQA$_{\text{zh}}$ & 136k/5k/3k & Chinese \\
\bottomrule
\end{tabular}
\caption{Summary of the evaluation benchmarks.}
\label{tbl:datasets}
\end{table}

We summarize all the evaluation benchmarks in Table~\ref{tbl:datasets}.
The datasets cover two tasks, and two more languages.
We report ROUGE~\cite{lin-2004-rouge} scores as the evaluation metrics for abstractive summarization. In addition, we include BLEU~\cite{bleu} and METEOR~\cite{meteor} metrics for question generation.

\subsubsection{Monolingual Dataset}

\noindent
\textbf{CNN / DailyMail}~\cite{see-etal-2017-get}
The abstractive summarization dataset aims at generating a concise and fluent summary from an English news article crawled from CNN and DailyMail. 

\noindent
\textbf{XSum}~\cite{xsum-emnlp}
The extreme summarization dataset compresses a BBC news article to a one-sentence summary.

\noindent
\textbf{SQuAD}~\cite{du-qg-2018} 
The question generation dataset aims at generating relevant questions given a paragraph and an answer span, which is based on SQuAD v1.1~\cite{squad1}.

\subsubsection{Multilingual Dataset}

\noindent
\textbf{Gigaword$_{\text{fr/zh}}$}~\cite{xnlg}
The headline generation datasets are built upon French (fr) and Chinese (zh) article-headline pairs.

\noindent
\textbf{WebQA$_{\text{zh}}$}~\cite{xnlg}
The Chinese question generation dataset is built upon WebQA~\cite{webqa}.

\subsection{Comparison of Fine-Tuning Methods}

\begin{table}[t]
\centering
\scalebox{0.8}{
\begin{tabular}{lcc}
\toprule
{\bf Method} & {\bf XSum} & {\bf SQuAD} \\
 & RG-1/RG-2/RG-L & BLEU-4/MTR/RG-L \\ \midrule
Causal & 40.72/18.44/33.30 & 23.42/25.07/49.96 \\
Masked & \textbf{41.12}/18.52/33.51 & 23.53/25.19/51.00 \\
Pseudo-Masked & 41.04/\textbf{18.69}/\textbf{33.58} & \textbf{23.61}/\textbf{25.36}/\textbf{51.05} \\
\bottomrule
\end{tabular}
}
\caption{Results of different fine-tuning methods on the XSum and SQuAD development sets.
The models are initialized with the BERT-base-uncased checkpoint.
RG is short for ROUGE, MTR for METEOR.}
\label{tbl:exp:methods}
\end{table}

We first compare the three sequence-to-sequence fine-tuning algorithms using the BERT-base-uncased checkpoint\footnote{\url{github.com/google-research/bert}} as the pretrained model.
We report evaluation results on the developments sets of XSum and SQuAD in Table~\ref{tbl:exp:methods}.
The results show that pseudo-masked fine-tuning achieves the best performance on two datasets, except that masked fine-tuning obtains the highest ROUGE-1 score on XSum.
Moreover, causal fine-tuning is consistently worse than the other two algorithms.
The results indicate that reducing the discrepancy between masked language model pre-training and sequence-to-sequence fine-tuning is beneficial.
We therefore use pseudo-masked fine-tuning in the rest of the experiments.

\begin{table}[t]
\centering
\scalebox{0.75}{
\begin{tabular}{lcc}
\toprule
{\bf Pretrained Model} & {\bf XSum} & {\bf SQuAD} \\
 & RG-1/RG-2/RG-L & BLEU-4/MTR/RG-L \\
\midrule
ELECTRA & 40.65/18.03/33.23 & 21.24/23.65/49.39 \\
BERT & 41.04/18.69/33.58 & 23.61/25.36/51.05 \\
RoBERTa & 43.30/20.47/35.53 & 25.32/26.61/52.62 \\
UniLMv2 & \textbf{44.45}/\textbf{21.67}/\textbf{36.78} & \textbf{26.30}/\textbf{27.09}/\textbf{53.19} \\
\bottomrule
\end{tabular}
}
\caption{
Evaluation results of four pretrained bidirectional Transformers on the development sets of XSum and SQuAD.
Pseudo-masked fine-tuning is used. The models are all base size.
Same shorthands apply as in Table~\ref{tbl:exp:methods}. 
}
\label{tbl:exp:models}
\end{table}

\subsection{Comparison of Pretrained Models}

\begin{table*}[t]
\centering
\scalebox{1.05}{
\small
\begin{tabular}{lllcc}
\toprule
{\bf Model} & {\bf \#Param} & {\bf Corpus} & {\bf CNN / DailyMail} & {\bf XSum} \\
& & & { RG-1/RG-2/RG-L} & { RG-1/RG-2/RG-L} \\ \midrule
\multicolumn{5}{l}{~~\textit{Without pre-training}} \\
\textsc{PtrNet}~\cite{see-etal-2017-get} & - & - & 39.53/17.28/36.38  & 28.10/8.02/21.72 \\ \midrule
\multicolumn{5}{l}{~~\textit{Fine-tuning base-size pretrained models}} \\
{MASS}~\cite{mass} & 123M & - &  42.12/19.50/39.01 & 39.75/17.24/31.95 \\
\textsc{BERTSumAbs}~\cite{bertsum} & 156M & 16GB &  41.72/19.39/38.76 & 38.76/16.33/31.15 \\
{\textsc{ERNIE-GEN}}~\cite{erniegen} & 110M & 16GB &  42.30/19.92/39.68 & -  \\
{T5}~\cite{t5} & 220M & 750GB &  42.05/20.34/39.40 & - \\
\textbf{\stosft}$_{\text{RoBERTa-base}}$ & 125M & 160GB & 42.28/20.21/39.87 & 43.39/20.55/35.63 \\
\textbf{\stosft}$_{\text{UniLMv2-base}}$ & 110M & 160GB & \textbf{43.89}/\textbf{21.05}/\textbf{41.02} & \textbf{44.37}/\textbf{21.54}/\textbf{36.61} \\
\midrule
\multicolumn{5}{l}{~~\textit{Fine-tuning {large}-size pretrained models}} \\
{UniLM}~\cite{unilm} & 340M & 16GB &  43.08/20.43/40.34 & - \\
{\textsc{ERNIE-GEN}}~\cite{erniegen} & 340M & 16GB &  44.02/21.17/41.26 \\
{BART}~\cite{bart} & 400M & 160GB & 44.16/21.28/40.90 & 45.14/22.27/37.25 \\
{ProphetNet}~\cite{pnet} & 400M & 160GB & 44.20/21.17/41.30 & - \\
{PEGASUS$_{\textsc{C4}}$}~\cite{PEGASUS} & 568M & 750GB & 43.90/21.20/40.76 & 45.20/22.06/36.99 \\
{PEGASUS$_{\textsc{HugeNews}}$}~\cite{PEGASUS} & 568M & 3800GB & 44.17/21.47/41.11 & 47.21/\textbf{24.56}/39.25 \\
{T5$_{\textsc{11B}}$}~\cite{t5} & 11B & 750GB & 43.52/21.55/40.69 & - \\
\textbf{\stosft}$_{\text{RoBERTa-large}}$ & 355M & 160GB & 43.92/21.25/41.06  & 45.63/22.72/37.86 \\
\textbf{\stosft}$_{\text{UniLMv2-large}}$ & 340M & 160GB & \textbf{44.79}/\textbf{21.98}/\textbf{41.93}  & \textbf{47.58}/24.35/\textbf{39.50} \\
\bottomrule
\end{tabular}
}
\caption{
Abstractive summarization results on the test set of CNN / DailyMail, and XSum.
The evaluation metric is the F1 version of ROUGE (RG) scores.
We also present the number of parameters (\#Param) for the methods using pretrained models.
}
\label{tbl:exp:summ}
\end{table*}

We compare different pretrained models for initialization, including BERT~\cite{bert}, ELECTRA~\cite{electra}, RoBERTa~\cite{roberta} and UniLMv2~\cite{unilmv2}.
The base-size checkpoints are used in the comparison.
As shown in Table~\ref{tbl:exp:models}, we report the results of pseudo-masked fine-tuning (Section~\ref{sec:pseudo:ft}) on XSum and SQuAD.

Among the four pretrained models, UniLMv2 performs best in terms of the automatic evaluation metrics, which contains a partially autoregressive pre-training objective that is similar to sequence-to-sequence modeling.
The models initialized by BERT and RoBERTa obtain better results compared with ELECTRA.
The results indicate that masked language model pre-training over the full vocabulary are helpful for sequence-to-sequence tasks. Although ELECTRA obtains comparable performance on a wide range of language understanding tasks (e.g., text classification, and question answering), the language modeling ability is not fully pretrained~\cite{electra}.

\subsection{Comparisons with Previous Work}

We conduct evaluation by using \stosft{} to fine-tune RoBERTa~\cite{roberta} and UniLMv2~\cite{unilmv2} on abstractive summarization (i.e., \cnndm{}, and XSum) and question generation (i.e., SQuAD).
Pseudo-masked fine-tuning is used both base-size and large-size models.

\begin{table*}[t]
\centering
\scalebox{1.07}{
\small
\begin{tabular}{lllcc}
\toprule
{\bf Model} & {\bf \#Param} & {\bf Corpus} & {\bf Official Split} & {\bf Reversed Split} \\
& & & { BLEU-4/MTR/RG-L} & { BLEU-4/MTR/RG-L} \\ \midrule
\multicolumn{5}{l}{~~\textit{Without pre-training}} \\
\cite{du-qg-2018} & - & - & 15.16/19.12/\hspace*{1.0 em}-\hspace*{1.0 em} & - \\
\cite{zhao-qg-2018} & - & - & - & 16.38/20.25/44.48 \\
\cite{zhang-qg-2019} & - & - & 18.37/22.65/46.68  & 20.76/24.20/48.91 \\ \midrule

\multicolumn{5}{l}{~~\textit{Fine-tuning {base}-size pretrained models}} \\
{\textsc{ERNIE-GEN}}~\cite{erniegen} & 110M & 16GB & 22.28/25.13/50.58  & 23.52/25.61/51.45 \\
\textbf{\stosft}$_{\text{RoBERTa-BASE}}$   & 125M  &160GB  & 23.86/25.93/51.68 & 25.32/26.61/52.62 \\
\textbf{\stosft}$_{\text{UniLMv2-BASE}}$   & 110M  &160GB  & \textbf{24.70}/\textbf{26.33}/\textbf{52.13}  & \textbf{26.30}/\textbf{27.09}/\textbf{53.19} \\ \midrule
\multicolumn{5}{l}{~~\textit{Fine-tuning {large}-size pretrained models}} \\
{UniLM}~\cite{unilm} & 340M  & 16GB  & 22.12/25.06/51.07  & 23.75/25.61/52.04 \\
{\textsc{ERNIE-GEN}}~\cite{erniegen} & 340M & 16GB & 24.03/26.31/52.36  & 25.57/26.89/53.31 \\
{ProphetNet}~\cite{pnet} & 400M & 16GB & 25.01/26.83/52.57  & 26.72/27.64/53.79  \\
\textbf{\stosft}$_{\text{RoBERTa-LARGE}}$ & 400M & 160GB & 25.30/26.85/52.66  & 26.82/27.48/53.92 \\
\textbf{\stosft}$_{\text{UniLMv2-LARGE}}$ & 340M & 160GB & \textbf{25.97}/\textbf{27.33}/\textbf{53.43}  & \textbf{27.12}/\textbf{27.95}/\textbf{54.25} \\
\bottomrule
\end{tabular}
}
\caption{
Question generation results on the test set of SQuAD. MTR is short for METEOR, and RG for ROUGE.
The official split is from~\cite{du-qg-2018}, while the reversed split is the same as in~\cite{zhao-qg-2018}.
}
\label{tbl:exp:qg}
\end{table*}

As shown in Table~\ref{tbl:exp:summ} and Table~\ref{tbl:exp:qg}, {\stosft}$_{\text{UniLMv2}}$ achieves state-of-the-art performance on all three benchmarks compared with the models that use more parameters, larger corpus, or task-specific pre-training.
Specifically, {T5$_{\textsc{11B}}$}~\cite{t5} uses 11 billion parameters and 750GB text corpus to pretrain a sequence-to-sequence model.
{PEGASUS}~\cite{PEGASUS} is a task-specific pretrained model designed for abstractive summarization.
The comparisons indicate that \stosft{} can obtain strong performance on sequence-to-sequence tasks by leveraging the pretrained models.

It is notable that RoBERTa obtains very competitive performance compared with previous work.
The comparisons show that the masked language modeling pre-training~\cite{bert} is helpful for language generation tasks.
Moreover, \stosft{} provides a unified modeling method to employ the existing pretrained Transformers for sequence-to-sequence tasks.

\begin{table*}[t]
\centering
\scalebox{0.97}{
\begin{tabular}{lccccc}
\toprule
& \bf WebQA$_{\text{zh}}$ & \bf Gigaword$_{\text{zh}}$ & \bf Gigaword$_{\text{fr}}$ \\
 & (Chinese) & (Chinese) & (French) \\
& BLEU-4/MTR/RG-L & RG-1/RG-2/RG-L & RG-1/RG-2/RG-L & & \\
\midrule
XLM~\cite{xnlg} & 23.41/23.32/47.40 & 55.30/42.57/52.95 & 56.27/39.20/52.84 \\
XNLG~\cite{xnlg} & 24.89/24.53/49.72 &  57.65/44.93/54.95 & 57.84/40.81/54.24 \\
\textbf{\stosft}$_{\text{XLM-RoBERTa-BASE}}$ & 27.45/25.20/49.76 & 60.29/47.24/57.46 & 57.95/41.30/54.54 \\
\textbf{\stosft}$_{\text{XLM-RoBERTa-LARGE}}$ & \textbf{28.49}/\textbf{26.48}/\textbf{52.94} & \textbf{60.95}/\textbf{47.94}/\textbf{58.09} & \textbf{58.48}/\textbf{41.79}/\textbf{55.04}\\
\bottomrule
\end{tabular}
}
\caption{Evaluation results of Chinese and French abstractive summarization, and Chinese question generation. QG is short for question generation, AS for abstractive summarization, BL for BLEU, MTR for METEOR, and RG for ROUGE.
}
\label{tbl:exp:xlm}
\end{table*}

\subsection{Results of Multilingual Generation}

Apart from monolingual generation tasks, we can use \stosft{} to leverage the multilingual pretrained models, such as mBERT~\cite{bert}, and XLM-RoBERTa~\cite{xlmr}.
We conduct language generation experiments on both abstractive summarization (French Gigaword$_{\text{fr}}$, and Chinese Gigaword$_{\text{zh}}$) and Chinese question generation (WebQA$_{\text{zh}}$).

As shown in Table~\ref{tbl:exp:xlm}, we employ \stosft{} to fine-tune XLM-RoBERTa on the three benchmarks.
We compare our results with fine-tuning XNLG~\cite{xnlg} and XLM~\cite{xlm} that are pretrained conventional sequence-to-sequence Transformers.
\stosft{} achieves significantly better performance than previous work across different languages and tasks.
The results indicate that \stosft{} can unleash the multilinguality of XLM-RoBERTa on generation tasks.
More importantly, the support of multilingual pretrained models greatly widens the application range of our \stosft{} toolkit.

\section{Conclusion}

We introduce a sequence-to-sequence toolkit \stosft{} to fine-tune the pretrained bidirectional Transformers for language generation tasks.
The toolkit follows the UniLM~\cite{unilm,unilmv2} fine-tuning algorithms, which unifies encoding and decoding with the same modeling method.
We conduct extensive experiments on abstractive summarization and question generation, including both monolingual and multilingual settings.
We plug in different pretrained models in our toolkit and evaluate three fine-tuning approaches.
Then we compare \stosft{} with previous work using both base-size and large-size models.
In addition, we use \stosft{} to apply off-the-shelf multilingual pretrained model on Chinese and French sequence-to-sequence learning.
Experimental results show that the proposed toolkit achieves strong performance across the tasks and languages.
We believe the toolkit is important to unleash the abilities of BERT-like bidirectional Transformers on sequence-to-sequence tasks.

\bibliography{s2s_ft}

\begin{thebibliography}{33}
\expandafter\ifx\csname natexlab\endcsname\relax\def\natexlab#1{#1}\fi

\bibitem[{Banerjee and Lavie(2005)}]{meteor}
Satanjeev Banerjee and Alon Lavie. 2005.
\newblock \href {https://www.aclweb.org/anthology/W05-0909} {{METEOR}: {A}n
  automatic metric for {MT} evaluation with improved correlation with human
  judgments}.
\newblock In \emph{Proceedings of the {ACL} Workshop on Intrinsic and Extrinsic
  Evaluation Measures for Machine Translation and/or Summarization}, pages
  65--72, Ann Arbor, Michigan. Association for Computational Linguistics.

\bibitem[{Bao et~al.(2020)Bao, Dong, Wei, Wang, Yang, Liu, Wang, Gao, Piao,
  Zhou, and Hon}]{unilmv2}
Hangbo Bao, Li~Dong, Furu Wei, Wenhui Wang, Nan Yang, Xiaodong Liu, Yu~Wang,
  Jianfeng Gao, Songhao Piao, Ming Zhou, and Hsiao{-}Wuen Hon. 2020.
\newblock \href {http://proceedings.mlr.press/v119/bao20a.html} {Unilmv2:
  Pseudo-masked language models for unified language model pre-training}.
\newblock In \emph{Proceedings of the 37th International Conference on Machine
  Learning, {ICML} 2020, 13-18 July 2020, Virtual Event}, volume 119 of
  \emph{Proceedings of Machine Learning Research}, pages 642--652. {PMLR}.

\bibitem[{Chi et~al.(2020)Chi, Dong, Wei, Wang, Mao, and Huang}]{xnlg}
Zewen Chi, Li~Dong, Furu Wei, Wenhui Wang, Xian{-}Ling Mao, and Heyan Huang.
  2020.
\newblock \href {https://www.aaai.org/Papers/AAAI/2020GB/AAAI-ChiZ.7682.pdf}
  {Cross-lingual natural language generation via pre-training}.
\newblock In \emph{The Thirty-Fourth {AAAI} Conference on Artificial
  Intelligence, {AAAI} 2020, New York, NY, USA, February 7-12, 2020}, pages
  7570--7577. {AAAI} Press.

\bibitem[{Clark et~al.(2020)Clark, Luong, Le, and Manning}]{electra}
Kevin Clark, Minh-Thang Luong, Quoc~V. Le, and Christopher~D. Manning. 2020.
\newblock \href {https://openreview.net/pdf?id=r1xMH1BtvB} {{ELECTRA}:
  Pre-training text encoders as discriminators rather than generators}.
\newblock In \emph{ICLR}.

\bibitem[{Conneau et~al.(2020)Conneau, Khandelwal, Goyal, Chaudhary, Wenzek,
  Guzm{\'{a}}n, Grave, Ott, Zettlemoyer, and Stoyanov}]{xlmr}
Alexis Conneau, Kartikay Khandelwal, Naman Goyal, Vishrav Chaudhary, Guillaume
  Wenzek, Francisco Guzm{\'{a}}n, Edouard Grave, Myle Ott, Luke Zettlemoyer,
  and Veselin Stoyanov. 2020.
\newblock \href {https://doi.org/10.18653/v1/2020.acl-main.747} {Unsupervised
  cross-lingual representation learning at scale}.
\newblock In \emph{Proceedings of the 58th Annual Meeting of the Association
  for Computational Linguistics, {ACL} 2020, Online, July 5-10, 2020}, pages
  8440--8451. Association for Computational Linguistics.

\bibitem[{Conneau and Lample(2019)}]{xlm}
Alexis Conneau and Guillaume Lample. 2019.
\newblock \href
  {http://papers.nips.cc/paper/8928-cross-lingual-language-model-pretraining.pdf}
  {Cross-lingual language model pretraining}.
\newblock In \emph{Advances in Neural Information Processing Systems}, pages
  7057--7067. Curran Associates, Inc.

\bibitem[{Devlin et~al.(2019)Devlin, Chang, Lee, and Toutanova}]{bert}
Jacob Devlin, Ming{-}Wei Chang, Kenton Lee, and Kristina Toutanova. 2019.
\newblock \href {https://doi.org/10.18653/v1/n19-1423} {{BERT:} {P}re-training
  of deep bidirectional transformers for language understanding}.
\newblock In \emph{Proceedings of the 2019 Conference of the North American
  Chapter of the Association for Computational Linguistics: Human Language
  Technologies, {NAACL-HLT} 2019, Minneapolis, MN, USA, June 2-7, 2019, Volume
  1 (Long and Short Papers)}, pages 4171--4186. Association for Computational
  Linguistics.

\bibitem[{Dong et~al.(2019)Dong, Yang, Wang, Wei, Liu, Wang, Gao, Zhou, and
  Hon}]{unilm}
Li~Dong, Nan Yang, Wenhui Wang, Furu Wei, Xiaodong Liu, Yu~Wang, Jianfeng Gao,
  Ming Zhou, and Hsiao{-}Wuen Hon. 2019.
\newblock \href
  {http://papers.nips.cc/paper/9464-unified-language-model-pre-training-for-natural-language-understanding-and-generation}
  {Unified language model pre-training for natural language understanding and
  generation}.
\newblock In \emph{Advances in Neural Information Processing Systems 32: Annual
  Conference on Neural Information Processing Systems 2019, NeurIPS 2019, 8-14
  December 2019, Vancouver, BC, Canada}, pages 13042--13054.

\bibitem[{Du and Cardie(2018)}]{du-qg-2018}
Xinya Du and Claire Cardie. 2018.
\newblock \href {https://doi.org/10.18653/v1/P18-1177} {Harvesting
  paragraph-level question-answer pairs from wikipedia}.
\newblock In \emph{Proceedings of the 56th Annual Meeting of the Association
  for Computational Linguistics, {ACL} 2018, Melbourne, Australia, July 15-20,
  2018, Volume 1: Long Papers}, pages 1907--1917. Association for Computational
  Linguistics.

\bibitem[{Lewis et~al.(2020)Lewis, Liu, Goyal, Ghazvininejad, Mohamed, Levy,
  Stoyanov, and Zettlemoyer}]{bart}
Mike Lewis, Yinhan Liu, Naman Goyal, Marjan Ghazvininejad, Abdelrahman Mohamed,
  Omer Levy, Veselin Stoyanov, and Luke Zettlemoyer. 2020.
\newblock \href {https://doi.org/10.18653/v1/2020.acl-main.703} {{BART:}
  denoising sequence-to-sequence pre-training for natural language generation,
  translation, and comprehension}.
\newblock In \emph{Proceedings of the 58th Annual Meeting of the Association
  for Computational Linguistics, {ACL} 2020, Online, July 5-10, 2020}, pages
  7871--7880. Association for Computational Linguistics.

\bibitem[{Li et~al.(2016)Li, Li, He, Wang, Cao, Zhou, and Xu}]{webqa}
Peng Li, Wei Li, Zhengyan He, Xuguang Wang, Ying Cao, Jie Zhou, and Wei Xu.
  2016.
\newblock Dataset and neural recurrent sequence labeling model for open-domain
  factoid question answering.
\newblock \emph{arXiv preprint arXiv:1607.06275}.

\bibitem[{Lin(2004)}]{lin-2004-rouge}
Chin-Yew Lin. 2004.
\newblock \href {https://www.aclweb.org/anthology/W04-1013} {{ROUGE}: A package
  for automatic evaluation of summaries}.
\newblock In \emph{Text Summarization Branches Out: Proceedings of the {ACL}-04
  Workshop}, pages 74--81, Barcelona, Spain. Association for Computational
  Linguistics.

\bibitem[{Liu(2019)}]{bertsum}
Yang Liu. 2019.
\newblock \href {http://arxiv.org/abs/1903.10318} {Fine-tune {BERT} for
  extractive summarization}.
\newblock \emph{CoRR}, abs/1903.10318.

\bibitem[{Liu and Lapata(2019)}]{liu-text}
Yang Liu and Mirella Lapata. 2019.
\newblock \href {https://doi.org/10.18653/v1/D19-1387} {Text summarization with
  pretrained encoders}.
\newblock In \emph{Proceedings of the 2019 Conference on Empirical Methods in
  Natural Language Processing and the 9th International Joint Conference on
  Natural Language Processing (EMNLP-IJCNLP)}, pages 3728--3738, Hong Kong,
  China. Association for Computational Linguistics.

\bibitem[{Liu et~al.(2019)Liu, Ott, Goyal, Du, Joshi, Chen, Levy, Lewis,
  Zettlemoyer, and Stoyanov}]{roberta}
Yinhan Liu, Myle Ott, Naman Goyal, Jingfei Du, Mandar Joshi, Danqi Chen, Omer
  Levy, Mike Lewis, Luke Zettlemoyer, and Veselin Stoyanov. 2019.
\newblock \href {http://arxiv.org/abs/1907.11692} {Roberta: {A} robustly
  optimized {BERT} pretraining approach}.
\newblock \emph{CoRR}, abs/1907.11692.

\bibitem[{Narayan et~al.(2018)Narayan, Cohen, and Lapata}]{xsum-emnlp}
Shashi Narayan, Shay~B. Cohen, and Mirella Lapata. 2018.
\newblock \href {https://doi.org/10.18653/v1/d18-1206} {Don't give me the
  details, just the summary! topic-aware convolutional neural networks for
  extreme summarization}.
\newblock In \emph{Proceedings of the 2018 Conference on Empirical Methods in
  Natural Language Processing, Brussels, Belgium, October 31 - November 4,
  2018}, pages 1797--1807. Association for Computational Linguistics.

\bibitem[{Papineni et~al.(2002)Papineni, Roukos, Ward, and Zhu}]{bleu}
Kishore Papineni, Salim Roukos, Todd Ward, and Wei-Jing Zhu. 2002.
\newblock \href {https://doi.org/10.3115/1073083.1073135} {{BLEU}: {A} method
  for automatic evaluation of machine translation}.
\newblock In \emph{Proceedings of the 40th Annual Meeting of the Association
  for Computational Linguistics}, pages 311--318, Philadelphia, Pennsylvania,
  USA. Association for Computational Linguistics.

\bibitem[{Radford et~al.(2018)Radford, Narasimhan, Salimans, and
  Sutskever}]{gpt}
Alec Radford, Karthik Narasimhan, Tim Salimans, and Ilya Sutskever. 2018.
\newblock \href
  {https://s3-us-west-2.amazonaws.com/openaiassets/research-covers/language-unsupervised/language
  understanding paper.pdf} {Improving language understanding by generative
  pre-training}.

\bibitem[{Raffel et~al.(2019)Raffel, Shazeer, Roberts, Lee, Narang, Matena,
  Zhou, Li, and Liu}]{t5}
Colin Raffel, Noam Shazeer, Adam Roberts, Katherine Lee, Sharan Narang, Michael
  Matena, Yanqi Zhou, Wei Li, and Peter~J. Liu. 2019.
\newblock \href {http://arxiv.org/abs/1910.10683} {Exploring the limits of
  transfer learning with a unified text-to-text transformer}.
\newblock \emph{arXiv e-prints}.

\bibitem[{Rajpurkar et~al.(2016)Rajpurkar, Zhang, Lopyrev, and Liang}]{squad1}
Pranav Rajpurkar, Jian Zhang, Konstantin Lopyrev, and Percy Liang. 2016.
\newblock \href {https://doi.org/10.18653/v1/D16-1264} {{SQ}u{AD}: 100,000+
  questions for machine comprehension of text}.
\newblock In \emph{Proceedings of the 2016 Conference on Empirical Methods in
  Natural Language Processing}, pages 2383--2392, Austin, Texas. Association
  for Computational Linguistics.

\bibitem[{Rothe et~al.(2020)Rothe, Narayan, and Severyn}]{bertshared}
Sascha Rothe, Shashi Narayan, and Aliaksei Severyn. 2020.
\newblock \href {https://transacl.org/ojs/index.php/tacl/article/view/1849}
  {Leveraging pre-trained checkpoints for sequence generation tasks}.
\newblock \emph{Trans. Assoc. Comput. Linguistics}, 8:264--280.

\bibitem[{See et~al.(2017)See, Liu, and Manning}]{see-etal-2017-get}
Abigail See, Peter~J. Liu, and Christopher~D. Manning. 2017.
\newblock \href {https://doi.org/10.18653/v1/P17-1099} {Get to the point:
  Summarization with pointer-generator networks}.
\newblock In \emph{Proceedings of the 55th Annual Meeting of the Association
  for Computational Linguistics (Volume 1: Long Papers)}, pages 1073--1083,
  Vancouver, Canada. Association for Computational Linguistics.

\bibitem[{Song et~al.(2019)Song, Tan, Qin, Lu, and Liu}]{mass}
Kaitao Song, Xu~Tan, Tao Qin, Jianfeng Lu, and Tie{-}Yan Liu. 2019.
\newblock \href {http://proceedings.mlr.press/v97/song19d.html} {{MASS:} masked
  sequence to sequence pre-training for language generation}.
\newblock In \emph{Proceedings of the 36th International Conference on Machine
  Learning, {ICML} 2019, 9-15 June 2019, Long Beach, California, {USA}},
  volume~97 of \emph{Proceedings of Machine Learning Research}, pages
  5926--5936. {PMLR}.

\bibitem[{Taylor(1953)}]{taylor1953cloze}
Wilson~L Taylor. 1953.
\newblock Cloze procedure: A new tool for measuring readability.
\newblock \emph{Journalism Bulletin}, 30(4):415--433.

\bibitem[{Vaswani et~al.(2017)Vaswani, Shazeer, Parmar, Uszkoreit, Jones,
  Gomez, Kaiser, and Polosukhin}]{transformer}
Ashish Vaswani, Noam Shazeer, Niki Parmar, Jakob Uszkoreit, Llion Jones,
  Aidan~N Gomez, {\L}ukasz Kaiser, and Illia Polosukhin. 2017.
\newblock \href
  {http://papers.nips.cc/paper/7181-attention-is-all-you-need.pdf} {Attention
  is all you need}.
\newblock In \emph{Advances in Neural Information Processing Systems 30}, pages
  5998--6008. Curran Associates, Inc.

\bibitem[{Wang and Cho(2019)}]{bertmarkovlm}
Alex Wang and Kyunghyun Cho. 2019.
\newblock \href {http://arxiv.org/abs/1902.04094} {{BERT} has a mouth, and it
  must speak: {BERT} as a markov random field language model}.
\newblock \emph{CoRR}, abs/1902.04094.

\bibitem[{Wolf et~al.(2019)Wolf, Debut, Sanh, Chaumond, Delangue, Moi, Cistac,
  Rault, Louf, Funtowicz, and Brew}]{HuggingFacesTS}
Thomas Wolf, Lysandre Debut, Victor Sanh, Julien Chaumond, Clement Delangue,
  Anthony Moi, Pierric Cistac, Tim Rault, R'emi Louf, Morgan Funtowicz, and
  Jamie Brew. 2019.
\newblock {HuggingFace}'s {Transformers}: State-of-the-art natural language
  processing.
\newblock \emph{ArXiv}, abs/1910.03771.

\bibitem[{Xiao et~al.(2020)Xiao, Zhang, Li, Sun, Tian, Wu, and Wang}]{erniegen}
Dongling Xiao, Han Zhang, Yu{-}Kun Li, Yu~Sun, Hao Tian, Hua Wu, and Haifeng
  Wang. 2020.
\newblock \href {http://arxiv.org/abs/2001.11314} {{ERNIE-GEN:} {An} enhanced
  multi-flow pre-training and fine-tuning framework for natural language
  generation}.
\newblock \emph{CoRR}, abs/2001.11314.

\bibitem[{Yan et~al.(2020)Yan, Qi, Gong, Liu, Duan, Chen, Zhang, and
  Zhou}]{pnet}
Yu~Yan, Weizhen Qi, Yeyun Gong, Dayiheng Liu, Nan Duan, Jiusheng Chen, Ruofei
  Zhang, and Ming Zhou. 2020.
\newblock \href {http://arxiv.org/abs/2001.04063} {{ProphetNet}: Predicting
  future n-gram for sequence-to-sequence pre-training}.
\newblock \emph{CoRR}, abs/2001.04063.

\bibitem[{Yang et~al.(2019)Yang, Dai, Yang, Carbonell, Salakhutdinov, and
  Le}]{xlnet}
Zhilin Yang, Zihang Dai, Yiming Yang, Jaime~G. Carbonell, Ruslan Salakhutdinov,
  and Quoc~V. Le. 2019.
\newblock \href
  {http://papers.nips.cc/paper/8812-xlnet-generalized-autoregressive-pretraining-for-language-understanding}
  {{XLNet}: Generalized autoregressive pretraining for language understanding}.
\newblock In \emph{Advances in Neural Information Processing Systems 32: Annual
  Conference on Neural Information Processing Systems 2019, NeurIPS 2019, 8-14
  December 2019, Vancouver, BC, Canada}, pages 5754--5764.

\bibitem[{Zhang et~al.(2020)Zhang, Zhao, Saleh, and Liu}]{PEGASUS}
Jingqing Zhang, Yao Zhao, Mohammad Saleh, and Peter~J. Liu. 2020.
\newblock \href {http://proceedings.mlr.press/v119/zhang20ae.html} {{PEGASUS:}
  pre-training with extracted gap-sentences for abstractive summarization}.
\newblock In \emph{Proceedings of the 37th International Conference on Machine
  Learning, {ICML} 2020, 13-18 July 2020, Virtual Event}, volume 119 of
  \emph{Proceedings of Machine Learning Research}, pages 11328--11339. {PMLR}.

\bibitem[{Zhang and Bansal(2019)}]{zhang-qg-2019}
Shiyue Zhang and Mohit Bansal. 2019.
\newblock \href {https://doi.org/10.18653/v1/D19-1253} {Addressing semantic
  drift in question generation for semi-supervised question answering}.
\newblock In \emph{Proceedings of the 2019 Conference on Empirical Methods in
  Natural Language Processing and the 9th International Joint Conference on
  Natural Language Processing, {EMNLP-IJCNLP} 2019, Hong Kong, China, November
  3-7, 2019}, pages 2495--2509. Association for Computational Linguistics.

\bibitem[{Zhao et~al.(2018)Zhao, Ni, Ding, and Ke}]{zhao-qg-2018}
Yao Zhao, Xiaochuan Ni, Yuanyuan Ding, and Qifa Ke. 2018.
\newblock \href {https://www.aclweb.org/anthology/D18-1424} {Paragraph-level
  neural question generation with maxout pointer and gated self-attention
  networks}.
\newblock In \emph{Proceedings of the 2018 Conference on Empirical Methods in
  Natural Language Processing}, pages 3901--3910, Brussels, Belgium.
  Association for Computational Linguistics.

\end{thebibliography}
\bibliographystyle{acl_natbib}

\newpage
\appendix

\section{Hyperparameters for Fine-Tuning}

Table~\ref{tbl:fxied_hyper} reports the most hyperparameters used in this paper. 

\begin{table}[htbp]
\centering
\begin{tabular}{lc}
\toprule
Batch size & 64 \\
Label smoothing & 0.1 \\
Adam $\epsilon$ & 1e-6 \\
Adam $\beta$ & (0.9, 0.999) \\
Learning rate schedule & Linear \\
Warmup steps & 1000 \\
Gradient clipping & 1.0 \\
Dropout & 0.1 \\
Weight decay & 0.01 \\
\bottomrule
\end{tabular}
\caption{
Hyperparameters for fine-tuning.
}
\label{tbl:fxied_hyper}
\end{table}
The optimal hyperparameter values are task-specific and we provide a range of possible values that work well for various downstream tasks:
\begin{itemize}
    \item {\bf Learning rate for base-sized models}: 5e-5, 7e-5, 1e-4
    \item {\bf Learning rate for large-sized models}: 1e-5, 1.5e-5, 2e-5, 3e-5
    \item {\bf Number of fine-tuning epochs}: 10, 15, 20, 30
    \item {\bf Mask prob for target sequence}: 40\%, 50\%, 60\%, 70\%
\end{itemize}
Mask prob for target sequence denotes the probability that each token in target sequence is masked. We conduct grid search on the development sets to find the best hyperparameters and use for the test sets. The other task-specific hyperparameters are listed in Table~\ref{tbl:task_specific}.

\begin{table}[htbp]
\scalebox{0.7}{
\centering
\setlength{\tabcolsep}{1pt}
\begin{tabular}{lrrrrr}
\toprule
\bf Task &\bf \makecell{Max input \\ tokens} & \bf \makecell{Max output \\ tokens} & \bf \makecell{Beam \\ size} & \bf \makecell{Length \\ penalty} & \bf \makecell{Min output \\ tokens} \\
\midrule
CNN / DailyMail & 608 & 160 & 5 & 0.9 & 48 \\
XSum                   & 720 & 48  & 8 & 0.7 & 1  \\
SQuAD QG               & 384 & 32  & 8 & 1.3 & 5  \\
WebQA$_{\text{zh}}$ QG & 384 & 32  & 8 & 1.3 & 5  \\
Gigaword$_{\text{fr}}$ & 96  & 48  & 5 & 0.9 & 1  \\
\bottomrule
\end{tabular}
}
\caption{
Task-specific hyperparameters for evaluation benchmarks. 
}

\label{tbl:task_specific}
\end{table}

\end{document}